\documentclass{article}

    \usepackage[final,nonatbib]{neurips_2020}

\usepackage[utf8]{inputenc} % allow utf-8 input
\usepackage[T1]{fontenc}    % use 8-bit T1 fonts
\usepackage{hyperref}       % hyperlinks
\usepackage{url}            % simple URL typesetting
\usepackage{booktabs}       % professional-quality tables
\usepackage{amsmath,amsfonts,amssymb}      % blackboard math symbols
\usepackage{nicefrac}       % compact symbols for 1/2, etc.
\usepackage{microtype}      % microtypography
\usepackage{graphicx}
\usepackage{xcolor}
\usepackage{multicol}
\usepackage{algorithm}
\usepackage[noend]{algpseudocode}

\usepackage{subfigure}
\usepackage{booktabs} % for professional tables
\usepackage{algorithm}
\usepackage[noend]{algpseudocode}
\usepackage[english]{babel}
\usepackage{xr}
\usepackage[squaren]{SIunits}
\newtheorem{theorem}{Theorem}

\title{Tomographic Auto-Encoder: \\ Unsupervised Bayesian Recovery of Corrupted Data}

\author{%
  Francesco Tonolini \\
  University of Glasgow\\
  Glasgow, UK \\
  \texttt{2402432t@student.gla.ac.uk} \\
  \And
  Pablo G. Moreno \\
  Amazon Research\\
  London, UK \\
  \texttt{morepabl@amazon.com} \\
  \And
  Andreas Damianou \\
  Amazon Research\\
  Cambridge, UK \\
  \texttt{damianou@amazon.com} \\
  \And
  Roderick Murray-Smith \\
  University of Glasgow\\
  Glasgow, UK \\
  \texttt{roderick.murray-smith@glasgow.ac.uk} \\
}

\begin{document}

\maketitle

\begin{abstract}
 We propose a new probabilistic method for unsupervised recovery of corrupted data. Given a large ensemble of degraded samples, our method recovers accurate posteriors of clean values, allowing the exploration of the manifold of possible reconstructed data and hence characterising the underlying uncertainty. In this setting, direct application of classical variational methods often gives rise to collapsed densities that do not adequately explore the solution space. Instead, we derive our novel \textit{reduced entropy condition} approximate inference method that results in rich posteriors. We test our model in a data recovery task under the common setting of missing values and noise, demonstrating superior performance to existing variational methods for imputation and de-noising with different real data sets. We further show higher classification accuracy after imputation, proving the advantage of propagating uncertainty to downstream tasks with our model.
\end{abstract}

\section{Introduction}

Data sets are rarely clean and ready to use when first collected. More often than not, they need to undergo some form of pre-processing before analysis, involving expert human supervision and manual adjustments \cite{PRE1,PRE2}. Filling missing entries, correcting noisy samples, filtering collection artefacts and other similar tasks are some of the most costly and time consuming stages in the data modeling process and pose an enormous obstacle to machine learning at scale \cite{TIME}. 
% Data needs to go through these often manual adjustments before learning, effectively rendering fully automated systems unfeasible. 
Traditional data cleaning methods rely on some degree of supervision in the form of a clean dataset or some knowledge collected from domain experts. However, the exponential increase of the data collection and storage rates in recent years, makes any supervised algorithm impractical in the context of modern applications that consume millions or billions of datapoints.
In this paper, we introduce a novel variational framework to perform automated data cleaning and recovery without any example of clean data or prior signal assumptions.

\begin{figure}[t]
  \centering
  \vspace{-0.5cm}
  \includegraphics[width=\linewidth]{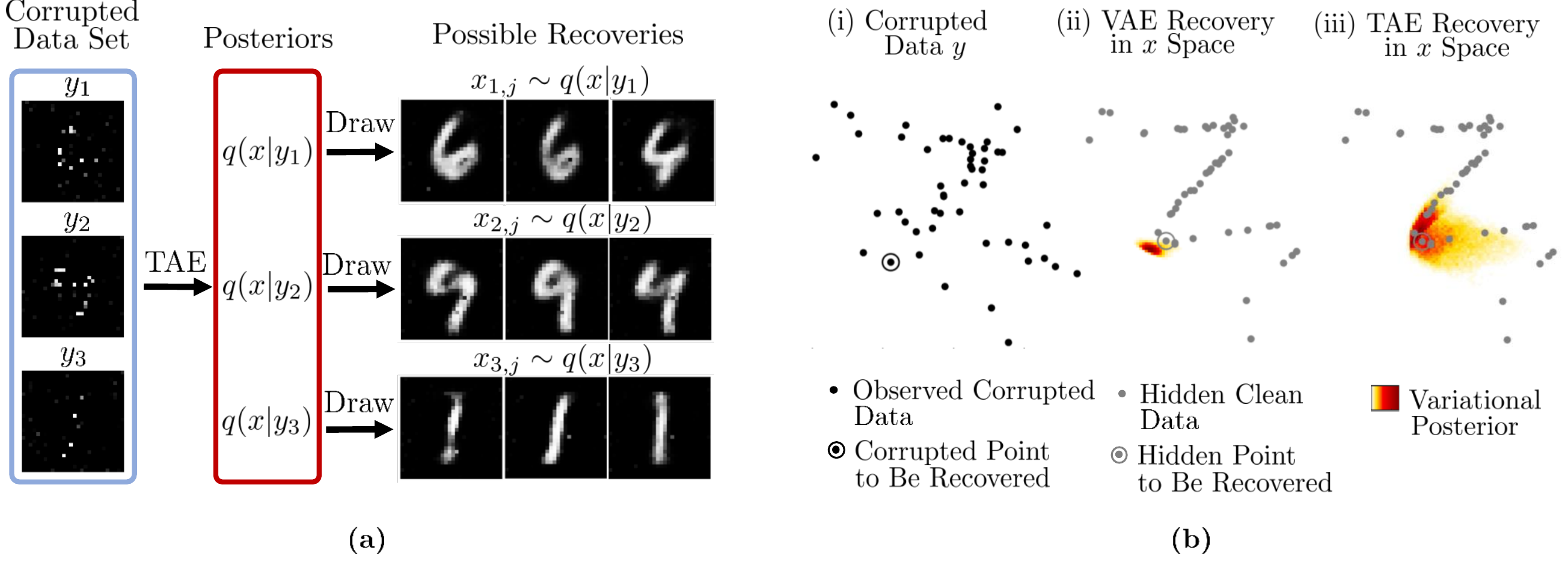}
\vspace{-0.3cm}
\caption{\textbf{(a)} Example of Bayesian recovery from corrupted data with a Tomographic Auto-Encoder (TAE) on corrupted MNIST. The TAE recovers posterior probability densities $q(x|y_i)$ for each corrupted sample $y_i$. We can draw from these to explore different possible clean solutions. \textbf{(b)} Two dimensional Bayesian recovery experiment. (i) Observed set of corrupted data $Y$, with the point we are inferring from $y_i$ highlighted. (ii) Ground truth hidden clean data with the target point $x_i$ highlighted, along with the posterior $q(x|y_i)$ reconstructed by a VAE. (iii) Posterior $q(x|y_i)$ recovered with our TAE. While the VAE posterior collapses to a single point, the TAE reconstructs a rich posterior that adjusts to the data manifold.}
\vspace{-0.3cm}
\label{fig:concept}
\end{figure}

The Tomographic auto-encoder (TAE), is named in analogy with standard tomography. 
Tomographic techniques for signal recovery aim at reconstructing a target signal, such as a 3D image, by algorithmically combining different incomplete measurements, such as 2D images from different view points, subsets of image pixels or other projections \cite{TOM1}. The TAE extends this concept to the reconstruction of data manifolds; our target signal is a clean data set, where corrupted data is interpreted as incomplete measurements. Our aim is to combine these to reconstruct the clean data set.

More specifically, we are interested in performing Bayesian recovery, where we do not simply transform degraded samples into clean ones, but recover probabilistic functions, with which we can generate diverse clean signals and capture uncertainty. Uncertainty is considerably important when cleaning data. If we are over-confident about specific solutions, errors are easily ignored and passed on to downstream tasks. For instance, in the example of figure \ref{fig:concept}(a), some corrupted observations are consistent with multiple digits. If we were to impute a single possibility for each sample, the true underlying solution may be ignored early on in the modeling pipeline and the digit will be consistently mis-classified. If we are instead able to recover accurate probability densities, we can remain adequately uncertain in any subsequent processing task. 

Several variational auto-encoder (VAE) models have been proposed for applications that can be considered special cases of this problem \cite{DEN_VAE1,MIS_VAE1,MUL_VAE1} and, in principle, they are capable of performing Bayesian reconstruction. 
However, we show that surrogating variational inference (VI) in a latent space with VAEs results in collapsed distributions that do not explore the different possibilities of clean samples, but only return single estimates. The TAE performs approximate VI in the space of recovered data instead, through our \textit{reduced entropy condition} method. The resulting posteriors adequately explore the manifold of possible clean samples for each corrupted observation and, therefore, adequately capture the uncertainty of the task.

% However, we demonstrate how directly applying VAEs to these recovery tasks often leads to collapsed distributions that do not explore the different possibilities of clean samples, but only return single estimates. We therefore propose a new inference algorithm which results in posteriors that deal with the ill-posedness of the recovery task, allowing exploration of the manifold of possible clean samples for each corrupted observation and, therefore, capture uncertainty. A challenge in implementing our inference model is the computation of the LVM entropy. We derive a novel framework to achieve this efficiently specifically suited for the data recovery task.

In our experiments we focus on data recovery from noisy samples and missing entries. This is one of the most common data corruption settings being encountered in a wide range of domains with different types of data \cite{MICE,MIS1}. By testing our approach in this prevalent scenario, we can closely compare with recently proposed VAE approaches \cite{MIS_VAE1, MIS_VAE2,MIS_VAE3}. We show how the existing VAE models exhibit the posterior collapse problem while the TAE produces rich posteriors that capture the underlying uncertainty. We further test TAEs on classification subsequent to imputation, 
demonstrating superior performance to existing methods in these downstream tasks.
% showing the advantage over existing methods in downstream tasks.
Finally, we use a TAE to perform automated missing values imputation on raw depth maps from the NYU rooms data set.

\section{Method}

We wish to build and train a parametric probability density function (PDF) $q(x|y)$, which takes as inputs corrupted samples $y$ and generates different possible corresponding clean data $x \sim q(x|y)$ through sampling. As natural data often lie on highly non-linear manifolds, we need this PDF to capture complicated modalities, e.g. the distribution of plausible images consistent with one of the corrupted observations in figure \ref{fig:concept}(a) explores diverse digits arranged in complex configurations in a high-dimensional space. A suitable recovery PDF $q(x|y)$ needs to be able to capture such complexity.

\begin{figure}
  \includegraphics[width=\textwidth]{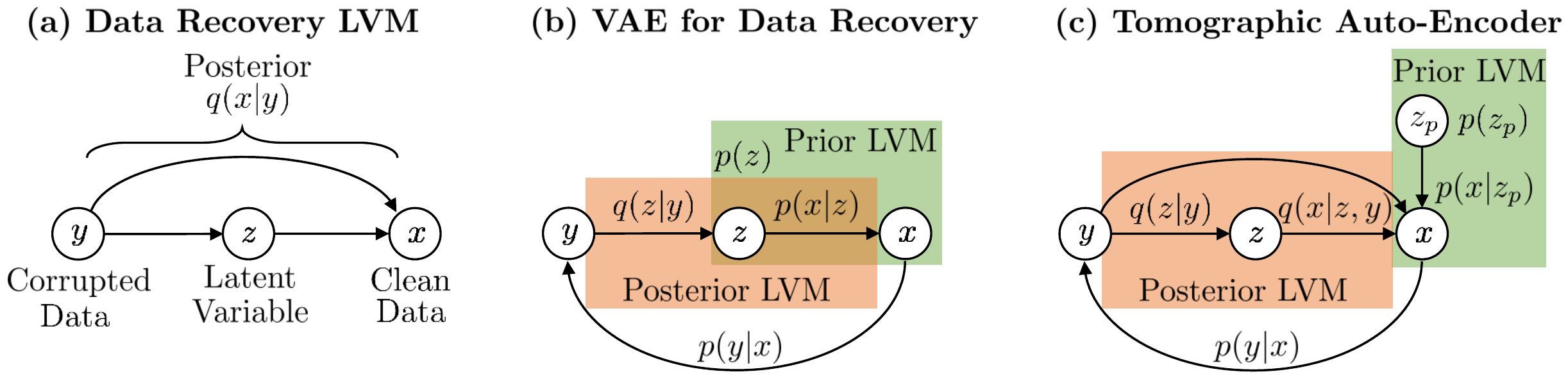}
  \vspace{-0.3cm}
  \caption{Training LVMs for data recovery. \textbf{(a)} Structure of a conditional LVM used to infer approximate posteriors $q(x|y)$ of clean data $x$ from corrupted observations $y$ as conditional inputs. \textbf{(b)} Training of $q(x|y)$ using a VAE. The clean data prior $p(x)$ and the approximate posterior $q(x|y)$ share the same latent space $z$ and generator $p(x|z)$, which results in collapsed posteriors. \textbf{(c)} Training of $q(x|y)$ using our TAE model. The prior $p(x)$ and posterior $q(x|y)$ are modelled with distinct LVMs, conceptually separating clean data structure from conditional inference.}
  \label{fig:models}
\end{figure}

In order to obtain distributions of sufficient capacity for the task, we construct $q(x|y)$ as a conditional latent variable model (LVM). Conditional LVM neural networks have achieved efficient and expressive variational inference in many recovery settings, capturing complex solution spaces in high dimensional problems, such as image reconstruction \cite{PLUG, CGAN, AH}. The conditional LVM consists of a first conditional distribution $q(z|y)$ mapping input corrupted data $y$ to latent variables $z$, and a second inference $q(x|z,y)$ mapping latent variables to output clean data $x$. The resulting PDF can be written
\begin{equation}\label{lvm_post}
\begin{split}
q(x|y) = \int q(z|y) q(x|z,y) dz,
\end{split}
\end{equation}
where both $q(z|y)$ and $q(x|z,y)$ are simple distributions, such as isotropic Gaussians, whose moments are inferred by neural networks taking the respective conditional arguments as inputs. Figure \ref{fig:models}(a) shows a graphical model for the conditional LVM.

% Given a corrupted data set $Y=\{ y_{1:N} \}$ and a corruption likelihood $p(y|x)$, we wish to fit a parametric approximation $q(x|y)$ to the posterior $p(x|y) \propto p(x)p(y|x)$, i.e, the distributions of possible clean data $x$ for each corrupted observation $y$. The prior $p(x)$ captures the structure of the clean data we want to reconstruct.
% In the general case, we can expect this to be rather complicated, as natural data often lies on highly non linear manifolds. In order to obtain posteriors of sufficient capacity for the task, we construct $q(x|y)$ as a conditional LVM. Conditional LVMs built with neural networks have achieved efficient and expressive variational inference in many recovery settings, capturing complex solution spaces in high dimensional problems, such as image reconstruction \citep{PLUG, CGAN, AH}. The conditional LVM consists in a first conditional distribution $q(z|y)$ mapping input corrupted data $y$ to latent variables $z$, and a second inference $q(x|z,y)$ mapping latent variables to output clean data $x$. Formally, we can write the resulting variational posterior as
% \begin{equation}\label{lvm_post}
% \begin{split}
% q(x|y) = \int q(z|y) q(x|z,y) dz,
% \end{split}
% \end{equation}
% where both $q(z|y)$ and $q(x|z,y)$ are simple distributions, such as isotropic Gaussians, whose moments are inferred by neural networks taking the respective conditional arguments as inputs. Figure \ref{fig:models}(a) shows a graphical model for the LVM posterior.

The approach to train $q(x|y)$ varies depending on the prior modelling assumptions we make and the data we have. For instance, in supervised variational inference, a training set of paired corrupted and clean data is used as training inputs $y$ and outputs $x$. The observed distributions of clean data $x$ can then be matched by parametric ones through a VAE or GAN training strategy \cite{CVAE,AH,VICI}.
% a training data set of paired clean and corrupted data represents samples from the joint distribution $p(x,y) = p(x)p(y|x)$, and therefore training a conditional LVM with such data is equivalent to fitting it to the true posterior $p(x|y)$  \citep{AH,VICI}. 

We are instead interested in the unsupervised situation, where we only have corrupted data $Y=\{ y_{1:N} \}$, a functional form for the corrupted data likelihood $p(y|x)$, e.g., missing values and additive noise, and we are interested in capturing the posterior $p(x|y)$, i.e, the distributions of possible clean data $x$ for each corrupted observation $y$. 
% but we have no example of clean data $x$.
Training a conditional LVM to fit posteriors without any ground truth examples $x$ is rather challenging, as we do not have data to encode from, in the case of VAE architectures, or adversarially compare with, in the case of GAN models. 

% Given a corrupted data set $Y=\{ y_{1:N} \}$ and a functional form for the corruption process likelihood $p(y|x)$, the aim is to recover a variational approximation to the posteriors $p(x|y)$; the distributions of possible clean data $x$ for each corrupted observation $y$. An LVM offers the possibility to jointly learn a data prior $p(x) = \int p(z)p_{\theta}(x|z)$ and fit a corresponding approximate posterior $q_{\phi}(x|y)$ to recover clean samples. However, as the prior over the signals of interest $p(x)$ is being learned from the observed data set $Y$, the capacity given to the model needs to be appropriately limited to avoid over-fitting. This is exactly the problem encountered when directly applying VAEs to data recovery, which we explain in detail below.

\subsection{VAEs and the Posterior Collapse Problem}

% \begin{figure}[t]
%   \centering
%   \includegraphics[width=\linewidth]{toy_1.pdf}
% \vspace{-0.7cm}
% \caption{Two dimensional Bayesian recovery experiment. \textbf{(a)} Observed set of corrupted data $Y$, with the point we are inferring from $y_i$ highlighted. \textbf{(b)} Ground truth hidden clean data with the target point $x_i$ highlighted, along with the posterior $q(x|y_i)$ reconstructed by a VAE. \textbf{(c)} Posterior $q(x|y_i)$ recovered with our TAE. While the VAE posterior collapses to a single point, the TAE reconstructs a rich posterior that adjusts to the data manifold.}
% \vspace{-0.5cm}
% \label{fig:toy}
% \end{figure}

Variational auto-encoders (VAEs) have been proposed for several problems within this definition of unsupervised reconstruction \cite{MIS_VAE2,DEN_VAE1,MUL_VAE1}. These methods lead to good single estimates of the underlying targets. However, they easily over-fit their posteriors resulting in collapsed PDFs $q(x|y)$. Put differently, they are often unable to explore different possible solutions to the recovery problem and return single estimates instead. Figure \ref{fig:concept}(b-ii) shows this pathology in a two dimensional experiment. 

The reason for this can be explained considering the structure of a VAE when modelling corrupted data and the resulting objective optimised when training the reconstruction posterior $q(x|y)$. The VAE encodes latent vectors $z$
 from corrupted observations $y$ with a conditional distribution $q(z|y)$ and reconstructs clean data $x$ with a decoding conditional $q(x|z)$. Reconstructed clean samples $x$ are then mapped back to corrupted samples $y$ with a corruption process likelihood $p(y|x)$, e.g. zeroing out missing entries, to maximise reconstruction of the observations $y$. Concurrently, regularisation in the latent space is induced with a user defined prior $p(z)$ (e.g. a unit Gaussian). The resulting lower bound to be maximised during training can be expressed as follows:
\begin{equation}\label{VAE_ELBO}
\begin{split}
\mathcal{L}_{VAE} = \mathbb{E}_{q(z|y)} \log p(y|z) - KL(q(z|y)||p(z)),
\end{split}
\end{equation}
where the observations likelihood is $p(y|z) = \int p(x|z)p(y|x)dx$ and in some cases, such as for missing values and additive noise, it is analytical.

When using a VAE formulation for unsupervised recovery, target data $x$ is completely hidden and, therefore, the model is implicitly introducing a signal prior $p(x) = \int p(z)p(x|z)dz$, which exploits the same decoder as the reconstruction posterior $\int q(z|x)p(x|z) dz$. This is shown in figure \ref{fig:models}(b). This leads to variational inference to only occur in $z$, where the KL divergence is tractable, rather than the target space $x$, resulting in the lower bound of equation \ref{VAE_ELBO} (see derivation in supplementary section A.1). While this may be computationally desirable, if $p(x|z)$ is of sufficient capacity, the model can learn to collapse distributions in $z$ to single estimates in $x$, failing to capture uncertainty. In fact, this is induced by the objective function of equation \ref{VAE_ELBO}; the model finds broad distributions in the latent space $q(z|y)$, which maximise the KL divergence, but the generator $p(x|z)$ collapses them back to single maximum likelihood solutions in x, maximising $\mathbb{E}_{q(z|y)} \log p(y|z)$.

\subsection{Separating Posterior and Prior: The Tomographic Auto-Encoder}

The proposed TAE sees instead independent models for the variational posterior and the prior, by adopting separate LVMs $q(x|y) = \int q(z|y) q(x|z,y) dz$ and $p(x) =  \int p(z_p) p(x|z_p) dz_p$ respectively (see figure \ref{fig:models}(c)). The de-coupling between prior and variational posterior results in variational inference occurring in clean data space $x$, instead of being surrogated to the latent space $z$ only, and we have freedom to design the signal prior LVM $p(x)$ as desired.

% allows control of the structure and capacity of the two, independently. 
% In fact, while empirical priors that are learned along with the variational posterior have been proven to perform well, their capacity normally needs to be carefully limited to avoid over-fitting \cite{VAMP, MG_VAE}. The TAE allows us to limit the capacity of the prior of clean data $p(x)$, while VAEs necessarily display matching generators for prior and variational posterior. 
The ELBO of the TAE model can be expressed as follows:
\begin{equation*}\label{TAE_ELBO}
\begin{split}
\mathcal{L}_{TAE} = \mathbb{E}_{q(x|y)} \log p(y|x) + \mathbb{E}_{q(x|y)} \big[\mathbb{E}_{q(z_p|x)} \log p(x|z_p)  
 - KL(q(z_p|x)||p(z_p)) \big] + H(q(x|y)).
\end{split}
\end{equation*}
The above ELBO is derived in detail in supplementary section A.2. The main difficulty here is to compute and maximise the self entropy of the approximate posterior $H(q(x|y))$, as this conditional distribution is an LVM of the form $q(x|y) = \int q(z|y)q(x|z,y) dz$.

\textbf{Reduced Entropy Condition:} Direct computation of the entropy of an LVM model $q(x|y) = \int_z q(z|y)q(x|z,y)dz$ is intractable in the general case. \cite{H_APP} proposed an approximate inference method to compute the gradient of the LVM's entropy for variational inference in latent spaces. However, this approach involves multiple samples to be drawn and evaluated with the LVM, which is expected to scale in complexity as the dimensionality and capacity of the target distribution increase.

In our case, we aim to approximately compute and optimise the entropy $H(q(x|y))$ for a distribution capturing natural data, which can be high-dimensional and lie on complicated manifolds. In order to maintain efficiency in the entropy estimation, we propose a new strategy; we identify a class of LVM posteriors for which the entropy reduces to a tractable form and then approximately constrain the posterior to such a class in our optimisation. Our main result is summarized in the following theorem:

\begin{theorem}\label{th}
If $\frac{q(z|x,y)}{q(z|y)} = B  \delta (z-g(x,y))$, where $\delta(\cdot)$ is the Dirac Delta function, $B$ is a real positive parameter and $g(x,y)$ is a deterministic function, then $H(q(x|y)) = H(q(z|y)) + \mathbb{E}_{q(z|y)} H(q(x|z,y))$.
\end{theorem}
% \textit{proof.} The proof is detailed in supplementary section A.3.

We detail the proof in supplementary A.3. Theorem \ref{th} states that if the posterior over latent variables $q(z|x,y)$ is infinitely more localised than the latent conditional $q(z|y)$, then the LVM entropy $H(q(x|y))$ has the tractable form given above. This condition imposes the LVM posterior to present non-overlapping conditionals $q(x|z,y)$ for different latent variables $z$, but does not impose any explicit restriction to the capacity of the model.
We can also formulate the condition as follows:
\begin{eqnarray}\label{cond}
\begin{split}
\mathbb{E}_{q(x,z|y)} \log \frac{q(z|x,y)}{q(z|y)} = C, \quad C \to \infty.
\end{split}
\end{eqnarray}

The proof is provided in supplementary section A.4. To train our posterior $q(x|y)$, we aim to maximise the ELBO $\mathcal{L}_{TAE}$ with the reduced entropy, while enforcing the condition of equation \ref{cond}:
\begin{equation}\label{TAE_OBJ}
\begin{split}
\arg \max \quad & \mathbb{E}_{q(x|y)} \log p(y|x) + \mathbb{E}_{q(x|y)} \big[\mathbb{E}_{q(z_p|x)} \log p(x|z_p)  - KL(q(z_p|x)||p(z_p)) \big] \\ + H(q(z|y)) &+ \mathbb{E}_{q(z|y)} H(q(x|z,y)),
 \quad s.t. \quad \mathbb{E}_{q(x,z|y)} \log \frac{q(z|x,y)}{q(z|y)} = C, \quad C \to \infty.
\end{split}
\end{equation}
While the ELBO is now amenable to stochastic optimization, the constraint is intractable since $C \to \infty$ and the posterior $q(z|x,y)$ is intractable.

\textbf{Relaxed Constraint:} To render the constraint tractable, we firstly relax $C$ to be a positive hyper-parameter. The higher the value of $C$, the more localised $q(z|x,y)$ is imposed to be compared to $q(z|y)$ and the closest the reduced entropy is to the true one.

To address the intractability of the posterior $q(z|x,y)$, we employ a variational approximation with a parametric function $r(z|x,y)$. In fact, for any valid probability density $r(z|x,y)$, we can prove that
\begin{eqnarray}\label{var_rel_lb}
\begin{split}
&\mathbb{E}_{q(x,z|y)} \log \frac{q(z|x,y)}{q(z|y)} \geq \mathbb{E}_{q(x,z|y)} \log \frac{r(z|x,y)}{q(z|y)}.
\end{split}
\end{eqnarray}
The proof is given in supplementary section A.5. The above bound implicates the following:
\begin{eqnarray*}\label{equiv}
\begin{split}
\mathbb{E}_{q(x,z|y)} \log \frac{r(z|x,y)}{q(z|y)} = C \equiv \mathbb{E}_{q(x,z|y)} \log \frac{q(z|x,y)}{q(z|y)} \geq C.
\end{split}
\end{eqnarray*}
This means that imposing the condition with a parametric distribution $r(z|x,y)$, which is trained along with the rest of the model, ensures deviation from the set condition only by excess. As the exact condition is met only at $\mathbb{E}_{q(x,z|y)} \log \frac{q(z|x)}{q(z|y)} \to \infty$, we can never relax the constraint more than already set by the finite value of $C$.

\textbf{The TAE Objective Function:} Having defined a tractable ELBO and a tractable condition, we need to perform the constrained optimisation
\begin{equation}\label{TAE_OBJ_2}
\begin{split}
\arg \max \quad &\mathbb{E}_{q(x|y)} \log p(y|x) + \mathbb{E}_{q(x|y)} \big[\mathbb{E}_{q(z_p|x)} \log p(x|z_p)  - KL(q(z_p|x)||p(z_p)) \big] \\ + &H(q(z|y)) + \mathbb{E}_{q(z|y)} H(q(x|z,y)),
    \quad s.t. \quad \mathbb{E}_{q(x,z|y)} \log \frac{r(z|x,y)}{q(z|y)} = C.
\end{split}
\end{equation}
We use the commonly adopted penalty function method \cite{PF,MUT_VAE} and relax equation \ref{TAE_OBJ_2} to an unconstrained optimisation with the use of a positive hyper-parameter $\lambda$:
\begin{equation}\label{TAE_OBJ_3}
\begin{split}
\arg \max \quad &\mathbb{E}_{q(x|y)} \log p(y|x) + \mathbb{E}_{q(x|y)} \big[\mathbb{E}_{q(z_p|x)} \log p(x|z_p)  - KL(q(z_p|x)||p(z_p)) \big] \\ &+ H(q(z|y)) + \mathbb{E}_{q(z|y)} H(q(x|z,y)) - \lambda \left| \mathbb{E}_{q(z,x|y)} \log \frac{r(z|x,y)}{q(z|y)} - C \right|.
\end{split}
\end{equation}
To train the model, we perform the maximisation of equation \ref{TAE_OBJ_3} using the ADAM optimiser. Once the model is trained, we can generate diverse reconstructions from a corrupt observation $y_i$ by sampling from the posterior $q(x|y_i)$. Details of our optimisation are reported in supplementary B.1. We describe how we handle parameters of the corruption process $p(y|x)$ in supplementary B.2.

% \textbf{Observation Parameters:} In the general case, the corruption process $p(y|x)$, mapping clean data $x$ to degraded samples $y$, is controlled by parameters that differ from sample to sample. We can distinguish these into observed parameters $\alpha$ and unobserved parameters $\beta$. For example, in the case of missing values and noise, the indexes of missing entries in each sample are often observed parameters, while the noise level is an unobserved parameter. The complete form of the corruption likelihood for a clean sample $x_i$ is then $p(y|x_i,\alpha_i,\beta_i)$. 

% In the inference, observed parameters $\alpha$ are additional conditional inputs analogous to degraded data $y$, giving posteriors $q(x|y,\alpha)$. Unobserved parameters $\beta$ need to be inferred from the observations themselves. We do this by inferring from the posterior's latent space with an additional parametric conditional $q(\beta|z,y)$, trained with the rest of the model. The objective function including these parameters is reported in supplementary section B.2. We also provide a pseudo-code in supplementary section B.3. Details on the choice of $C$ and $\lambda$ are given in supplementary D.1.

\section{Related Work}

\subsection{Supervised Bayesian Reconstruction}

The reconstruction of posterior densities from incomplete measurements has been recently investigated in supervised situations, where examples of clean data are available. In particular, conditional generative models were demonstrated with high dimensional data \cite{ITRA}. These methods work by exploiting an LVM to generate diverse realisations of targets conditioned on associated observations \cite{CGAN2,PLUG}.  Both conditional generative adversarial networks (CGANs) \cite{CGAN,CGAN2} and conditional VAEs (CVAEs) \cite{CVAE,PLUG} have been studied in this context. 
% In both cases, the samples generated by conditioning on an observation can be interpreted as samples from the corresponding conditional posterior densities. 

These approaches proved successful in a range of recovery tasks, such as reconstruction of images with missing groups of pixels \cite{PLUG}, super-resolution \cite{ITRA}, medical computed tomography reconstructions \cite{AH} and semi supervised situations, where examples of clean data and conditions are available in different amounts \cite{SEMI1,VICI}. Other works aim at reconstructing manifolds of solutions from observations only, but can still be considered supervised as they exploit pre-trained generators \cite{PRET_GAN}.

These works make the important observation that when learning to recover data from corrupted or partial observations, there is not a single right solution, but many  differently likely ones. We aim to extend this ability to completely unsupervised scenarios, where no clean data examples are available.

\subsection{Unsupervised Bayesian Reconstruction}

Reconstructing posteriors in the unsupervised case is largely still an open problem. However, several tasks that fall within this definition have been recently approached with Bayesian machine learning methods. Arguably the most investigated is de-noising.
% , i.e., given a noisy data set alone, we wish to train a model to return clean samples.
Several works solve this problem by exploiting the natural tendency of neural networks to regularise outputs \cite{N2N,N2V,PN2V}. Other methods build LVMs that explicitly model the noise process in their decoder, retrieving clean samples upon encoding and generation \cite{DEN_VAE1,DEN_VAE2}.

A second notable example is that of missing value imputation. Corrupted data corresponds to samples with missing entries. Recent works have explored the use of LVMs to perform imputation, both with GANs \cite{MI_GAN, MI_GAN2} and VAEs \cite{MIS_VAE1,MIS_VAE3}. In the former, the discriminator of the GAN is trained to distinguish real values from imputed ones, such that the generator is induced to synthesise realistic imputations. In the latter, the encoder of a VAE maps incomplete samples to a latent space, to then generate complete samples.

Finally, Bayesian LVM methods have been used on other unsupervised tasks that can be cast as special cases of data recovery problems. Amongst these, we find Multi-view generation \cite{MV_GAN, MUL_VAE1}, where the target clean data includes all views for each samples, but the observed data only presents subsets. Blind source separation can also be cast as a recovery problem and has been approached with GANs and VAEs \cite{BSS_VAE,BSS_GAN}.

\begin{figure}
  \vspace{-0.5cm}
  \includegraphics[width=\textwidth]{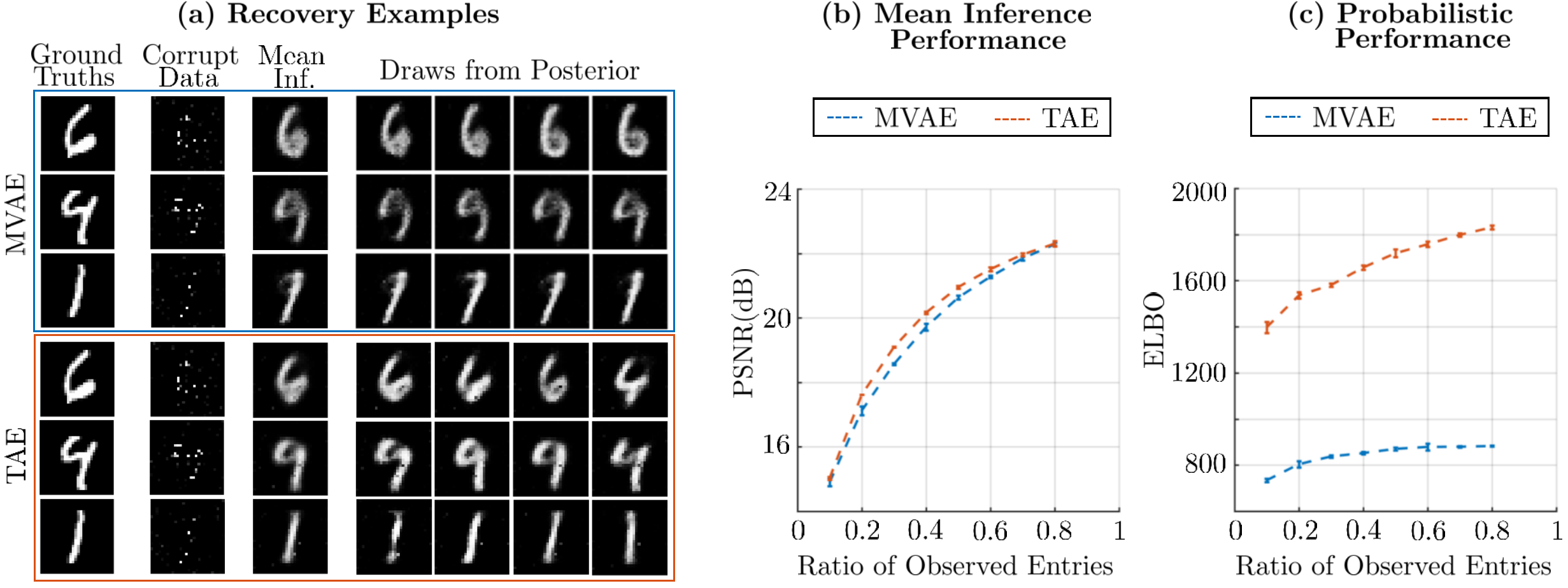}
  \vspace{-0.5cm}
  \caption{MNIST data recovery from missing entries and noise. \textbf{(a)} Recoveries using an MVAE and our TAE, showing average reconstruction and samples from the trained posteriors. \textbf{(b)} PSNR between ground truths and mean reconstruction. \textbf{(c)} ELBO assigned by the recovered posteriors to the ground truth data. The mean inference performance is very similar for the two models (PSNR values), while the probabilistic performance (ELBO values) is significantly higher for our TAE model. We can see evidence of this difference in the reconstruction examples; The MVAE and TAE return similarly adequate mean solutions, but the MVAE posterior's draws are all very similar, suggesting that the posterior has collapsed on a particular reconstruction. Contrarily, the posteriors returned by the TAE explore different possible solutions that are consistent with the associated corrupted observation.}
  \label{fig:MNIST}
\end{figure}

These models proved to be successful at reconstructing data in their specific domain.
However, in our work, we show how exploiting a standard VAE inference structure, similarly to several of the aforementioned methods, often leads to posteriors of clean data that collapse on single estimates, sacrificing the probabilistic capability of LVMs. 
% Instead, we introduce a specific variational posterior structure that gives expressive distributions when training on corrupted data, overcoming the posterior collapse pathology.

\section{Experiments}

% We evaluate TAE on several benchmark data sets with both missing entries and additive noise on the observed entries. Missing value and additive noise is arguably the most common degradation encountered in large data sets and allows us to directly compare TAEs with existing variational methods for imputation \cite{MICE,MIS_VAE1}.
\vspace{-0.1cm}
\subsection{Posterior Recovery}
We corrupt the MNIST dataset \cite{MNIST} by introducing missing values and additive Gaussian noise on the observed entry. We then train both a missing value imputation VAE (MVAE), analogous to those presented in \cite{MIS_VAE1} and \cite{MIS_VAE2}, and our TAE model with the corrupted data sets. The resulting variational posteriors are used to perform data recovery from the corrupted samples. Figure \ref{fig:MNIST}(a) shows, for different examples, the mean reconstruction and a set of draws from the posterior. We present analogous experiments for the missing-not-at-random case in supplementary D.2.

% \vspace{-0.3cm}
We evaluate the accuracy of mean reconstruction at different ratios of observed entries by measuring the peak signal to noise ratio (PSNR) between the ground truth data and mean recoveries (Figure \ref{fig:MNIST}(b)). To evaluate probabilistic performance we approximately measure the likelihood assigned by the recovered posteriors to the ground truth data through a reconstruction ELBO, by training a new inference function with the clean ground truths, but leaving the posterior fixed, as is common for evaluating ELBOs in unsupervised settings \cite{REFIT_1,REFIT_2,MIS_VAE3}. Results are shown in figure \ref{fig:MNIST}(c). We further evaluate our TAE with Fashion-MNIST -- $28 \times 28$ grey-scale images of clothing \cite{ZALANDO}, and the UCI HAR dataset, which consists of filtered accelerometer signals from mobile phones worn by different people during common activities \cite{HAR}. As before, we test the recovery of these data sets from a version affected by missing values and additive noise. In addition to the MVAE baseline, we also compared against the recently proposed missing values importance weighted auto encoder (MIWAE) \cite{MIS_VAE3}, which optimises an importance weighted ELBO in place of the standard one. For each model and settings we compute the ELBO assigned to the ground truth data. Results are shown in Table~\ref{tab:ELBO}. Experimental details in Sec.\ C of supplementary material.
\begin{center}
\begin{table}
\caption{Bayesian recovery from noisy data with different percentages of missing entries. Table shows the ELBO assigned by the retrieved posteriors to the ground truth clean data. Our TAE model consistently returns higher ELBO values compared to the competing variational methods, as it is able to retrieve rich posteriors that adequately sample the solution space. More values in supp. D.3.}
\label{tab:ELBO}
\vspace{0.2cm}
  \begin{tabular}{ l c c c c c c}
    \toprule
    & \multicolumn{2}{c}{MNIST}  & \multicolumn{2}{c}{Fashion-MNIST} & \multicolumn{2}{c}{UCI HAR} \\
      & $50\%$ & $80\%$ & $50\%$ & $80\%$ & $50\%$ & $80\%$ \\
    \hline
    MVAE & $870 \pm 6$ & $803 \pm 15$ & $757 \pm 1$ & $723 \pm 7$ & $585 \pm 4$ &  $471 \pm 10$ \\
    MIWAE & $917 \pm 4$ & $780 \pm 6$ & $800 \pm 7$ & $766 \pm 8$ & $613 \pm 6$ &  $584 \pm 8$ \\
    TAE & $\mathbf{1719 \pm 7}$ & $\mathbf{1536 \pm 14}$ & $\mathbf{1326 \pm 7}$ & $\mathbf{1094 \pm 13}$ & $\mathbf{1014 \pm 6}$ &  $\mathbf{854 \pm 52}$ \\
    % $20\%$ missing &   &  & \\ \hline
    % MNIST, & $870 \pm 6$ & $917 \pm 4$ & $\mathbf{1719 \pm 7}$ \\
    % $50\%$ missing &   &  & \\ \hline
    % MNIST, & $803 \pm 15$ & $780 \pm 6$ & $\mathbf{1536 \pm 14}$ \\
    % $80\%$ missing &   &  & \\ \hline
    % Fashion-MNIST, & $775 \pm 4$ & $815 \pm 4$ & $\mathbf{1407 \pm 24}$ \\
    % $20\%$ missing &   &  & \\ \hline
    % Fashion-MNIST, & $757 \pm 1$ & $800 \pm 7$ & $\mathbf{1326 \pm 7}$ \\
    % $50\%$ missing &   &  & \\ \hline
    % Fashion-MNIST, & $723 \pm 7$ & $766 \pm 8$ & $\mathbf{1094 \pm 13}$ \\
    % $80\%$ missing &   &  & \\ \hline
    % UCI HAR, & $611 \pm 3$ & $628 \pm 10$ & $\mathbf{1039 \pm 11}$ \\
    % $20\%$ missing &   &  & \\ \hline
    % UCI HAR, & $585 \pm 4$ & $613 \pm 6$ & $\mathbf{1014 \pm 6}$ \\
    % $50\%$ missing &   &  & \\ \hline
    % UCI HAR, & $471 \pm 10$ & $584 \pm 8$ & $\mathbf{854 \pm 52}$ \\
    % $80\%$ missing &   &  & \\ \bottomrule
  \end{tabular}
  \vspace{-0.4cm}
  \end{table}
\end{center}
\vspace{-0.7cm}
\subsection{Downstream Tasks}

To investigate the advantage of capturing complex uncertainties with our TAE model, we are interested in testing performance in downstream tasks. We test classification performance on subsets of the MNIST and Fashion-MNIST data sets, after recovery with our TAE. With both sets, we consider situations in which $10.000$ examples are available, but corrupted with missing entries and noise. $1,000$ of these are labelled with one of $10$ possible classes and we wish to classify the remaining $9,000$. To do so, we first train the TAE model on the full set, then use the recovered posteriors to generate multiple possible cleaned data for the labelled sub-set and use them to train a classifier.

\begin{figure}[htb!]
  \centering
%   \vspace{-0.7cm}
  \includegraphics[width=\linewidth]{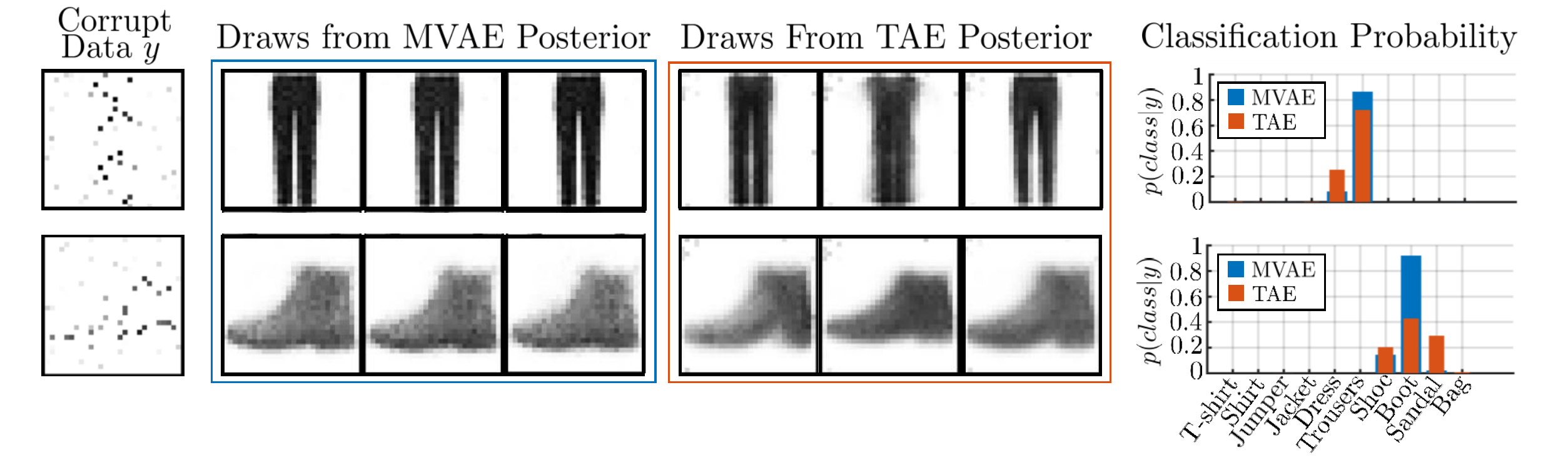}
\vspace{-0.6cm}
\caption{Propagating uncertainty to a classification task. Draws from the MVAE posterior are all very similar to each other. As a result, the imputed images are almost always classified in the same way and the uncertainty of the task is underestimated. The TAE posterior explores varied possible solutions to the recovery task. These can be recognised as different classes, resulting in less concentrated distributed probabilities that better reflect the associated uncertainty.}
% \vspace{-0.5cm}
\label{fig:class}
\end{figure}

\begin{figure}
  \centering
  \includegraphics[width=\linewidth]{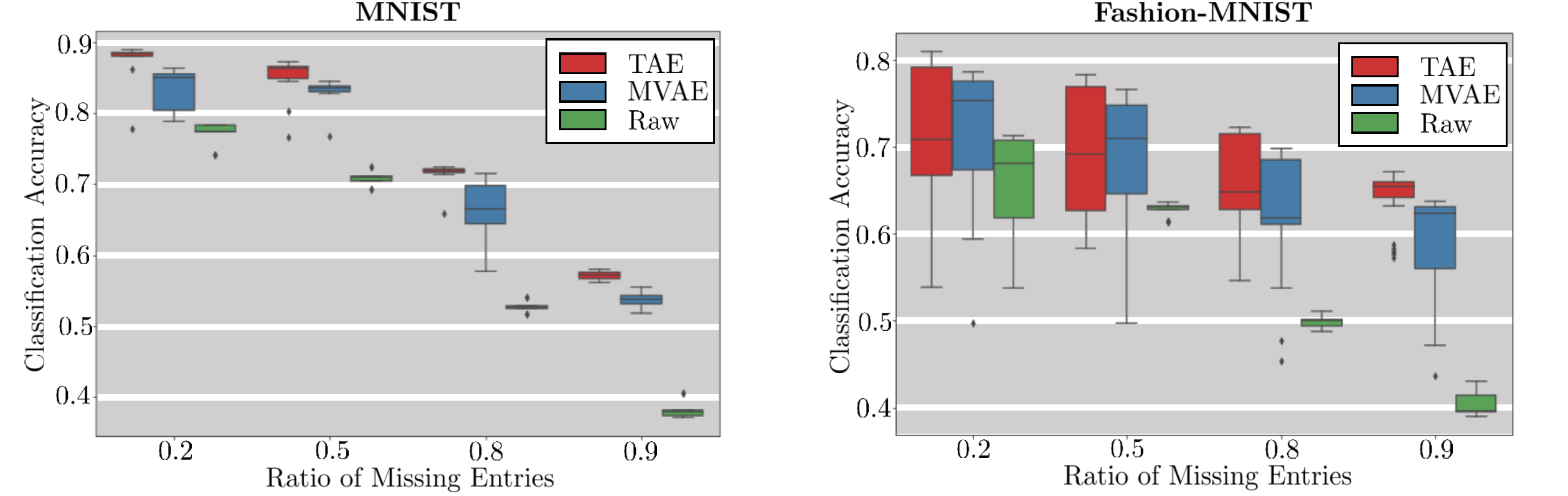}
\vspace{-0.4cm}
\caption{Classification accuracy after imputation. Classifying using TAE imputations gives an advantage in this downstream task over using raw corrupted data and MVAE imputations, especially when the number of missing entries is high. This is because the MVAE collapses on single imputations, while the TAE generates diverse samples for each corrupted observation. The TAE classifier trains with data augmentations consistent with observed corrupted images, instead of single estimates.}
\vspace{-0.5cm}
\label{fig:class_acc}
\end{figure}

To perform classification on the $9,000$ remaining examples, we  generate multiple possible cleaned data with the variational posteriors. Then, for each posterior sample, we perform classification and histogram the results. Examples are shown in figure \ref{fig:class}. To evaluate the performance, we take the class with the largest histogram as the inferred one. We repeat this experiment for different ratios of missing values and several repetitions, varying the subsets of labelled and unlabelled data to be used. Classification accuracy results are shown in figure \ref{fig:class_acc}.

\subsection{Missing Values in the NYU Depth Maps}

As a final practical application, we use a convolutional version of our TAE to perform structured missing value imputation on depth maps of indoors rooms collected with a Kinect depth sensor. Missing entries are very common in depth maps recorded with structured light sensors, such as the Kinect \cite{DM_1}. We use raw depth data from the NYU rooms dataset \cite{NYU}. This is composed of both RGB and matched depth images of indoors rooms. A small sub-set of the depth maps has been corrected by imputing the missing entries and is popularly employed to train and test various learning systems \cite{NYU_app_1,NYU_app_2,NYU_app_3}. However, a much larger portion of the set is available only as raw data, which presents missing entries. These are especially concentrated around objects' edges and reflecting surfaces, breaking the common assumption of missing at random, making this task particularly challenging.
\begin{figure}[htb]
  \centering
  \includegraphics[width=\linewidth]{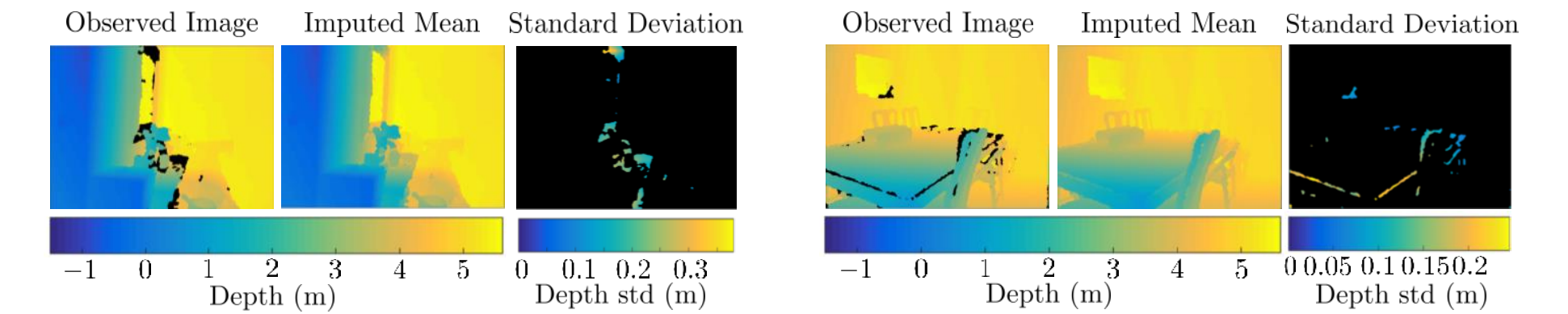}
\vspace{-0.4cm}
\caption{Unsupervised missing value imputation with our TAE on raw depth maps from the NYU rooms data set. Missing pixels in the raw images are in black. The TAE generates possibilities for the imputed pixels, which can be aggregated to recover a mean and standard deviation to quantify uncertainty in the retrieved imputations. More examples are shown in supplementary D.4.}
% \vspace{-0.7cm}
\label{fig:rooms}
\end{figure}

We train our TAE with a subset of this raw data set to perform imputation. In this case we are only interested in imputing missing entries and not denoising, as depth points that are registered by the Kinect present very low noise. Examples of results are shown in figure \ref{fig:rooms}. The imputation of this data set was carried out with no signal prior, no particular domain expertise, temporal or structural assumptions and no associated RGB images or examples of complete data. We simply run a convolutional version of the TAE on the raw depth maps to impute the missing values.

\vspace{-.1cm}
\section{Conclusion}
We presented tomographic auto-encoders; a variational inference method for recovering posterior distributions of clean data from a corrupted data set alone. We derive the \textit{reduced entropy condition} method; a novel inference strategy that results in rich distributions of clean data given corrupted observations, thereby capturing the uncertainty of the task, while standard variational methods often collapse on single answers. In our experiments, we demonstrate this capability and show the advantage of capturing uncertainty with the TAE in downstream tasks, outperforming the state-of-the-art VAE based recovery methods.

\newpage
\bibliography{example_paper} 
\bibliographystyle{ieeetr}

\newpage

{\Large\bf\centering Tomographic Auto-Encoder - Supplementary material \par} 

\appendix
\section{Proofs and Derivations}

\subsection{Derivation of VAE ELBO for Data Recovery}

We aim to maximise the log likelihood of the observed corrupted data $y$
\begin{equation}\label{supp:VAE_ELBO_1}
\begin{split}
\log p(y) = \log \int_x \underbrace{\int_z p(z)p(x|z) dz}_{p(x)} p(y|x) dx.
\end{split}
\end{equation}
We can introduce a variational distribution in both clean data space and latent space $q(x,z|y)$ and define a lower bound as
\begin{equation}\label{supp:VAE_ELBO_2}
\begin{split}
\log p(y) \geq  &\int_x \int_z q(x,z|y) \log \frac{p(z)p(x|z) dz}{q(x,z|y)} dz dx\\ + &\int_x \int_z q(x,z|y) \log p(y|x) dz dx.
\end{split}
\end{equation}
To obtain the VAE ELBO used in data recovery settings, the choice of the variational posterior is $q(x,z|y) = q(z|y)p(x|z)$. The ELBO can then be simplified to give
\begin{equation}\label{supp:VAE_ELBO_3}
\begin{split}
\log p(y) \geq  &\int_x \int_z q(z|y)p(x|z) \log \frac{p(z)p(x|z) dz}{q(z|y)p(x|z)} dz dx \\ + &\int_x \int_z q(z|y)p(x|z) \log p(y|x) dz dx \\ = & \underbrace{\int_x p(x|z) dx}_{=1} \int_z q(z|y) \log \frac{p(z) dz}{q(z|y)} dz \\ + &\int_x \int_z q(z|y)p(x|z) dz \log p(y|x) dx
\end{split}
\end{equation}
For situations in which the observations' likelihood $\int_x p(x|z)p(y|x) dx$ has a closed form, such as additive noise and missing entries, we can define a tighter bound to the likelihood by moving the integral in $x$ in the second term inside the logarithm
\begin{equation}\label{supp:VAE_ELBO_4}
\begin{split}
\log p(y) \geq & \int_z q(z|y) \log \frac{p(z) dz}{q(z|y)} dz \\ + & \int_z q(z|y) \log \left[\int_x p(y|x)p(x|z) dx \right] dz \\ =& -KL(q(z|y)||p(z)) + \mathbb{E}_{q(z|y)} \log p(y|z) dx.
\end{split}
\end{equation}
Because $p(x|z)$ simplifies in the KL term, this ELBO avoids variational inference in the space of clean data $x$.

\subsection{Derivation of TAE ELBO}

In our TAE model we defined separate LVMs for prior and posterior. To distinguish between the posterior latent variable and the prior latent variable, we name the former $z$ and the latter $z_p$. The likelihood we aim to maximise is
\begin{equation}\label{supp:TAE_ELBO_1}
\begin{split}
\log p(y) = \log \int_x \underbrace{\int_{z_p} p(z_p)p(x|z_p) dz_p}_{p(x)} p(y|x) dx.
\end{split}
\end{equation}
Similarly to the VAE ELBO case, we define a variational posterior $q(x,z_p|y)$ to find a lower bound
\begin{equation}\label{supp:TAE_ELBO_2}
\begin{split}
\log p(y) \geq  &\int_x \int_{z_p} q(x,z_p|y) \log \frac{p(z_p)p(x|z_p) dz_p}{q(x,z_p|y)} dz_p dx\\ + &\int_x \int_{z_p} q(x,z_p|y) \log p(y|x) dz_p dx.
\end{split}
\end{equation}
However, in our model we do not make the assumption that the variational posterior has the special form described in section A.1 and instead set it to have the form $q(x,z_p|y) = q(x|y)q(z_p|x)$, separating posterior inference from observations $y$ to clean data $x$ and inference of prior latent variables $z_p$. The resulting lower bound is
\begin{equation}\label{supp:TAE_ELBO_3}
\begin{split}
\log p(y) \geq  &\int_x \int_{z_p} q(x|y)q(z_p|x) \log \frac{p(z_p)p(x|z_p)}{q(x|y)q(z_p|x)} dz_p dx + \int_x \int_{z_p} q(x|y)q(z_p|x) \log p(y|x) dz_p dx \\ = & \int_x q(x|y) \underbrace{\int_{z_p} q(z_p|x) \log \frac{p(z_p)p(x|z_p)}{q(z_p|x)} dz_p}_{\geq \log p(x)} dx  + \int_x \underbrace{\int_{z_p} q(z_p|x)}_{=1} dz_p q(x|y) \log p(y|x) dx \\ - & \int_x \underbrace{\int_{z_p} q(z_p|x) dz_p}_{=1} q(x|y) \log q(x|y) dx \\ = & \mathbb{E}_{q(x|y)} \big[\mathbb{E}_{q(z_p|x)} \log p(x|z_p) - KL(q(z_p|x)||p(z_p)) \big] + \mathbb{E}_{q(x|y)} \log p(y|x) + H(q(x|y)) .
\end{split}
\end{equation}

\subsection{Proof of Theorem 1}

\begin{eqnarray}\label{del_id}
\begin{split}
&\frac{q(z|x,y)}{q(z|y)} = B \delta (z-g(x,y))
\implies \frac{q(z|x,y)}{q(z|y)} \frac{q(z'|x,y)}{q(z'|y)} = 0, \quad \forall x, z \neq z' \\ \implies &\frac{q(x|z,y)}{q(x|y)} \frac{q(x|z',y)}{q(x|y)} = 0, \quad \forall x, z \neq z' \\ \implies &q(x|z,y)q(x|z',y) = 0, \quad \forall x, z \neq z' \\ \implies &q(x|z',y) = 0, \quad \forall x \sim q(x|z,y), z \neq z'
\end{split}
\end{eqnarray}

Using the result of equation \ref{del_id}, we can derive the form of the entropy $H(q(x|y))$ for this special case as the following:

\begin{eqnarray}\label{hx}
\begin{split}
H(q(x|y)) = &-\int_x \left[ \int_z q(z|y) q(x|z,y) dz \right] \cdot \log \bigg[ \int_{z'} q(z'|y) q(x|z',y) dz' \bigg] dx \\ = &-\int_x \int_z q(z|y) q(x|z,y)  \cdot \log \bigg[ \int_{z'= z} q(z'|y) q(x|z',y) dz' \\ + &\underbrace{\int_{z'\neq z} q(z'|y) q(x|z',y) dz'}_{eq. \ref{del_id} \implies =0} \bigg] dz dx \\ &= -\int_z \int_x q(z|y) q(x|z,y) \log  \left[q(z|y) q(x|z,y) \right] dx dz \\ &=  -\int_z \int_x q(z|y) q(x|z,y) \log q(z|y) dx dz -\int_z \int_x q(z|y) q(x|z,y) \log q(x|z,y) dx dz \\ &= -\int_z q(z|y) \log q(z|y) dz - \int_z q(z|y) \int_x q(x|z,y) \log q(x|z,y) dx dz \\ &= H(q(z|y)) + \mathbb{E}_{q(z|y)} H(q(x|z,y)).
\end{split}
\end{eqnarray}

\subsection{Proof of the Equivalence Between Conditions}
\textbf{proof of necessary condition:}
\begin{eqnarray}\label{cond_eq_nec}
\begin{split}
\mathbb{E}_{q(x,z|y)} \log \frac{q(z|x,y)}{q(z|y)} = &\int_z q(z|y) \int_x q(x|z,y) \log \frac{q(z|x,y)}{q(z|y)} dx dz \\ = &\int_x q(x|y) \int_z q(z|x,y) \log \frac{q(z|x,y)}{q(z|y)} dz dx \\
= &\int_x q(x|y) \int_z q(z|x,y) \log q(z|x,y) dz dx \\ -&\int_x q(x|y) \int_z q(z|x,y) \log q(z|y) dz dx \\ =&\int_x q(x|y) \int_z q(z|x,y) \log q(z|x,y) dz dx \\ -& \underbrace{\int_x q(x|z,y)dx}_{=1} \int_z q(z|y) \log q(z|y) dz \\ = & \mathbb{E}_{q(x|y)} \underbrace{\int_z q(z|x,y) \log q(z|x,y) dz}_{-H(q(z|x,y))} \\ - &\underbrace{\int_z q(z|y) \log q(z|y) dz}_{-H(q(z|y))}.
\end{split}
\end{eqnarray}
If the above expression tends to infinity, either $H(q(z|x,y)) \to -\infty$ or $H(q(z|y)) \to \infty$, meaning that either $q(z|x,y) \to $ a Delta function, or $q(z|y) \to $ uniform. Either condition implies $\frac{q(z|x,y)}{q(z|y)} = B \delta (z-g(x,y))$.

\textbf{proof of sufficient condition:}
\begin{eqnarray}\label{cond_eq_suff}
\begin{split}
\mathbb{E}_{q(x,z|y)} \log \frac{q(z|x,y)}{q(z|y)} = &\int_x q(x|y) \int_z q(z|x,y) \log \frac{q(z|x,y)}{q(z|y)} dz dx \\ = &\int_x q(x|y) \int_z q(z|y) \frac{q(z|x,y)}{q(z|y)} \log \frac{q(z|x,y)}{q(z|y)} dz dx. 
\end{split}
\end{eqnarray}
Now we set $\frac{q(z|x,y)}{q(z|y)} = B \delta (z-g(x,y))$:
\begin{eqnarray}\label{cond_eq_suff_2}
\begin{split}
 &\int_x q(x|y) \int_z q(z|y) B \delta (z-g(x,y)) \log B \delta (z-g(x,y)) dz dx \\ = & \int_x q(x|y)  q(g(x,y)|y)\log B \underbrace{\delta (g(x,y)-g(x,y))}_{\to \infty, \forall x} dx.
\end{split}
\end{eqnarray}
Therefore, $\frac{q(z|x,y)}{q(z|y)} = B \delta (z-g(x,y))$ is a sufficient condition for $\mathbb{E}_{q(x,z|y)} \log \frac{q(z|x,y)}{q(z|y)} \to \infty$.

\subsection{Proof of Equation 6}

\begin{eqnarray}\label{supp_cond_1}
\begin{split}
\mathbb{E}_{q(x,z|y)} \log \frac{q(z|x,y)}{q(z|y)} = &\int_z \int_x q(x,z|y) \log q(z|x,y) dzdx \\ - &\int_z \int_x q(x,z|y) \log q(z|y) dzdx \\ = &\int_x q(x|y) \int_z q(z|x,y) \log q(z|x,y) dz dx \\ - &\int_z \int_x q(x,z|y) \log q(z|y) dzdx \\ \geq &\int_x q(x|y) \int_z q(z|x,y) \log r(z|x,y) dz dx \\ - &\int_z \int_x q(x,z|y) \log q(z|y) dzdx \\ = &\mathbb{E}_{q(x,z|y)} \log \frac{r(z|x,y)}{q(z|y)},
\end{split}
\end{eqnarray}
Where the inequality derives from the positivity of the KL divergence $KL(q(z|x,y)||r(z|x,y))$.

\section{Algorithm}

\subsection{Details of Training}

As detailed in section 3.2, to train our variational posterior $q(x|y)$, we maximise through gradient ascent the TAE ELBO with the reduced entropy penalty function
\begin{equation}\label{TAE_OBJ}
\begin{split}
\arg \max \quad &\mathbb{E}_{q(x|y)} \log p(y|x) + \mathbb{E}_{q(x|y)} \big[ \underbrace{\mathbb{E}_{q(z_p|x)} \log p(x|z_p)  - KL(q(z_p|x)||p(z_p))}_{Prior \quad  ELBO, \quad \geq p(x)} \big] \\ &+ H(q(z|y)) + \mathbb{E}_{q(z|y)} H(q(x|z,y)) - \lambda \left| \mathbb{E}_{q(z,x|y)} \log \frac{r(z|x,y)}{q(z|y)} - C \right|.
\end{split}
\end{equation}
All expectations in the above expression are computed and optimised by sampling the corresponding conditional distributions using the re-parametrisation trick characteristic of VAEs.

Because the prior LVM $p(x) = \int p(z_p)p(x|z_p) dz_p$ is training entirely with samples from the posterior LVM, which is also training, the model can easily obtain high values for the prior ELBO by generating collapsed samples $x$ with the posterior and get stuck in an unfavourable local minimum.  TO avoid this, we employ a warm up strategy. We define a positive parameter $\gamma$ that multiplies the expectation of the prior ELBO and the entropy $H(x|z,y)$:
\begin{equation}\label{TAE_WU}
\begin{split}
\arg \max \quad &\mathbb{E}_{q(x|y)} \log p(y|x) + \gamma \mathbb{E}_{q(x|y)} \big[ \underbrace{\mathbb{E}_{q(z_p|x)} \log p(x|z_p)  - KL(q(z_p|x)||p(z_p))}_{Prior \quad  ELBO, \quad \geq p(x)} \big] \\ &+ H(q(z|y)) + \gamma \mathbb{E}_{q(z|y)} H(q(x|z,y)) - \lambda \left| \mathbb{E}_{q(z,x|y)} \log \frac{r(z|x,y)}{q(z|y)} - C \right|.
\end{split}
\end{equation}
The value of $\gamma$ is initially set to zero. After a set number of iterations it is linearly increased to reach one and kept constant for the remaining training iterations.

\subsection{Complete Objective Function}

\textbf{Observation Parameters:} In the general case, the corruption process $p(y|x)$, mapping clean data $x$ to degraded samples $y$, is controlled by parameters that differ from sample to sample. We can distinguish these into observed parameters $\alpha$ and unobserved parameters $\beta$. For example, in the case of missing values and noise, the indexes of missing entries in each sample are often observed parameters, while the noise level is an unobserved parameter. The complete form of the corruption likelihood for a clean sample $x_i$ is then $p(y|x_i,\alpha_i,\beta_i)$. 

\textbf{Objective Function:} With the parameters conditionals described in subsection 3.2.4 and explicitly showing the parameters to be optimised, the objective function we maximise is the following
\begin{equation}\label{TAE_OBJ}
\begin{split}
\arg\max_{\theta,\phi} \quad &\mathbb{E}_{q_{\phi}(x,\beta|y,\alpha)} \log p(y|x,\alpha,\beta)\\ + &\gamma \mathbb{E}_{q_{\phi}(x|y,\alpha)} \big[ \mathbb{E}_{q_{\phi_3}(z_p|x)} \log p_{\theta}(x|z_p)  - KL(q_{\phi_3}(z_p|x)||p(z_p)) \big] \\ + &H(q_{\phi_1}(z|y,\alpha)) + \gamma \mathbb{E}_{q_{\phi_1}(z|y,\alpha)} H(q_{\phi_2}(x|z,y,\alpha)) \\ - &\lambda \left| \mathbb{E}_{q_{\phi}(z,x|y,\alpha)} \log \frac{r_{\phi_4}(z|x)}{q_{\phi_1}(z|y,\alpha)} - C \right|,
\end{split}
\end{equation}
$q_{\phi}(x,\beta|y,\alpha) = \int_z q_{\phi_1}(z|y,\alpha) q_{\phi_2}(x|z,y,\alpha) q_{\phi_5}(\beta|z,y,\alpha) dz$, $q_{\phi}(x|y,\alpha) = \int_z q_{\phi_1}(z|y,\alpha) q_{\phi_2}(x|z,y,\alpha) dz$,
$q_{\phi}(z,x|y,\alpha) = q_{\phi_1}(z|y,\alpha)q_{\phi_2}(x|z,y,\alpha)$,
$\phi = \{ \phi_{1:5} \}$ are the parameters of the inference models and $\theta$ are the parameter of the prior model.

\subsection{Pseudo-Code}

\begin{algorithm}[h!]
\caption{Training the TAE Model}\label{euclid}

\vspace{0.2cm}

\textbf{\textit{Inputs:}} Corrupted observations $Y = \{y_{1:N}\}$; Observed Parameters $A = \{\alpha_{1:N}\}$ initial model parameters, \{$\theta^{(0)}, \phi^{(0)}$\}; user-defined posterior latent dimensionality, $J$; user-defined prior latent dimensionality, $J_p$; user-defined condition strength $\lambda$; user-defined condition parameter $C$; user-defined latent prior $p(z_p)$; user-defined initial warm-up coefficient $\gamma_0$; user-defined final warm-up coefficient $\gamma_f$; warm-up start $N_{w0}$;  warm-up end $N_{wf}$; user-defined number of iterations, $N_{iter}$.
\vspace{0.2cm}
%
% \begin{adjustwidth}{}{}
\begin{algorithmic}[1]
\State $\gamma^{(k=0)} \gets \gamma_0$ 
%\Statex 
%\vspace{-0.2cm}
\State \textbf{for} \textit{the} $k$'th \textit{iteration} \textbf{in} $[0:N_{iter}-1]$ \\
\quad \textbf{for} \textit{the} $i$'th \textit{observation}  \\
\quad \quad $z_{i} \sim q_{\phi_1^{(k)}}(z|y_i,\alpha_i)$ \\
\quad \quad $x_{i} \sim q_{\phi_2^{(k)}}(x| z_i,y_i,\alpha_i)$ \\
\quad \quad $\beta_{i} \sim q_{\phi_5^{(k)}}(\beta| z_i,y_i,\alpha_i)$ \\
\quad \quad $z_{p,i} \sim q_{\phi_3^{(k)}}(z_p| x_i)$ \\
\quad \quad %\state
$\textbf{E}^{(k)}_i\gets \log p(y_i|x_i,\beta_i)$  \\
\quad \quad $\textbf{P}^{(k)}_i\gets \log p_{\theta^{(k)}}(x_i|z_{p,i})$  \\
\quad \quad $\textbf{K}^{(k)}_i \gets D_{KL}(q_{\phi_3^{(k)}}(z_p|x_i)|| p(z_p)$) \\
\quad \quad $\textbf{Hz}^{(k)}_i \gets H(q_{\phi_1^{(k)}}(z|y_i,\alpha_i))$ \\ \quad \quad $\textbf{Hx}^{(k)}_i \gets H(q_{\phi_2^{(k)}}(x|z_i,y_i,\alpha_i))$  \\ \quad \quad $\textbf{R}^{(k)}_i \gets \log r_{\phi_4^{(k)}}(z_i|x_i,y_i,\alpha_i)$  \\ \quad \quad $\textbf{Q}^{(k)}_i \gets \log q_{\phi_1^{(k)}}(z_i|y_i,\alpha_i)$  \\
\quad \textbf{end}  
\Statex \\
\quad $\textbf{F}^{(k)} = \sum_i\bigg( \textbf{E}^{(k)}_i + \gamma^{(k)} \left[ \textbf{P}^{(k)}_i - \textbf{K}^{(k)}_i + \textbf{Hx}^{(k)}_i  \right]$ \\  \quad \quad \quad $+ \textbf{Hz}^{(k)}_i - \lambda \left| \textbf{R}^{(k)}_i - \textbf{Q}^{(k)}_i - C \right|\bigg)$  \\ 
 \quad $\theta^{(k+1)}, \phi^{(k+1)} \gets {\arg\max}  (\textbf{F}^{(k)})$ 
 \Statex  \\ 
 \vspace{0.2cm}
 \quad \textbf{if} $k>N_{w0}$ \textbf{and} $k<N_{wf}$ \\
 \quad \quad $\gamma^{(k+1)} \gets \gamma^{(k)}+(\gamma_f-\gamma_0)/(N_{wf}-N_{w0})$ \\
 \quad \textbf{else} \\
  \quad \quad $\gamma^{(k+1)} \gets \gamma^{(k)}$ \\
 \quad \textbf{end} 
 \\ 
 \textbf{end}
 \Statex
\end{algorithmic}
% \end{adjustwidth}
\end{algorithm}

\section{Experimental Details}

\subsection{Models' Architectures}

In all experiments we carry out comparing our TAE with competitive methods, we make the independence assumption $q(x|z,y) = q(x|z)$, consequentially making $r(z|x,y) = r(z|x)$. In this way, the reconstruction posterior LVMs $q(x|y)$ we compare between TAE, MVAE and MIWAE all present identical structure and differences in performance are a result of the model constructed to train them alone. However, we note that, unlike the two competing method, the TAE is not formally limited to this choice and can infer conditionals $q(x|z,y)$ in the general case. We hereafter detail the architecture used for all quantitative experiments of section 4.1 and 4.2.

\textbf{Posteriors structure:} The posterior parametric components are $q_{\phi_1}(z|y,\alpha)$ and $q_{\phi_2}(x|z)$ ($p_{\phi_2}(x|z)$ in the case of the MVAE and MIWAE). $q_{\phi_1}(z|y,\alpha)$ consists in a fully connected two layers neural network with leaky ReLu non-linearities, taking as input concatenated corrupted observations $y$ and a binary mask that labels the missing entries $\alpha$ and returning as output a vector of latent means and a vector of latent log variances. The two intermediate deterministic layers have $400$ hidden units, while the latent space $z$ is $20$-dimensional.

$q_{\phi_2}(x|z)$, and $p_{\phi_2}(x|z)$ in the case of the MVAE and MIWAE, are similarly constructed, consisting in a fully connected two layers neural network with leaky ReLu non-linearities, taking as input latent variables $z$ and returning a vector of means and a vector of log variances of clean samples $x$. The two intermediate deterministic layers have $400$ hidden units.

\textbf{TAE Prior LVM Structure:} The TAE prior encoder $q_{\phi_3}(z_p|x)$ has the same general structure as the posterior encoder, with two fully connected layers and leaky ReLu non-linearities, taking as input generated clean data $x$ and returning as outputs a vector of latent means and a vector of latent log variances for the prior latent variable $z_p$. As this model has less capacity than the posterior LVM, the two deterministic hidden layers have $50$ hidden units each and the latent variables $z_p$ are $5$-dimensional.

$p_{\theta}(x|z_p)$ is similarly constructed, consisting in a fully connected two layers neural network with leaky ReLu non-linearities, taking as input latent variables $z_p$ and returning a vector of means and a vector of log variances of clean samples $x$. The two intermediate deterministic layers have $50$ hidden units.

\textbf{Approximate Latent Posterior Structure:} The approximate latent posterior $r(z|x)$ has the same structure as the posterior encoder, consisting in a fully connected two layers neural network with leaky ReLu non-linearities, taking as input generated clean data $x$ and returning as outputs a vector of latent means and a vector of latent log variances. The two intermediate deterministic layers have $400$ hidden units.

\textbf{Convolutional TAE Structure:} For the imputation of NYU missing data we use convolutional conditionals in our model, instead of fully connected ones. In this version. we do not make the independence assumption, using $q(x|z,y)$ and $r(z|x,y)$. $q(z|y,\alpha)$ takes concatenated $y$ and $\alpha$ and passes them through $4$ recurrent convolutional layers with filters of size $3 \times 3$ and 5 channels, each time down-sampling by two. the last layer is mapped to means and standard deviation of latent images $z$, which are $1/32$ of the original size in each axis and have $10$ channels, through two convolutional filter banks with strieds $1 \times 1$. $q(x|z,y, \alpha)$ is built to mirror this structure,  with the addition of accepting inputs from $y$ and $\alpha$. Three recurrent transpose convolutional layers with $3 \times 3$ filters, $5$ channels and $2 \times 2$ upsampling each map $z$ to a deterministic layer with $1/2 \times 1/2$ of the original images size. concatenated $y$ and $\alpha$ are mapped to the same size with a single convolutional layer, downsampling it by $1/2 \times 1/2$ and $5$ channels. The two are concatenated and the resulting layer is finally upsampled to inferred clean image $x$ by a last convolution with a filter bank. All non-linearities are Elu.

The prior networks are built in a similar way, but with shallower structures to give less capacity. $q(z_p|x)$ passes $x$ through $2$ convolutional layers, each with down-sampling of $4 \times 4$ and $5$ channels. as before, mens and standard deviations of latent images $z_p$ are generated from this last layers with $2 \times 2$ down-sampling and, in this case, $5$ channels. The prior generator $p(x|z_p)$ is built to exactly mirror this structure. $r(z|x,y,\alpha)$ has the same structure as $q(z|y,\alpha)$, with the only difference being that it accepts as input concatenated $x$, $y$ and $\alpha$.

\subsection{Experimental Conditions}

\textbf{Posterior Recovery:} All posterior recovery experiments, with each of the three data sets tested, are performed on samples that have been re-scaled from $0$ to $1$. In all cases, the sets are injected with additive Gaussian noise having standard deviation $0.1$. Subsequently, random binary masks are generated to block out some entries, resulting in missing values. The proportion of missing entries in the masks was set as described in the main body in each case.

Experiments were repeated with re-generated binary masks $5$ times. The means and error bars shown in figure 4 and the uncertainty reported in table 1 were computed from these. The MIWAE was trained with 20 weights per sample. After training, all posteriors $q(x|y)$ have identical structure and are tested in the same way, by training an inference network on the test set to compute the ELBO values.

\textbf{Classification Experiments:} The TAE models for the MNIST and Fashion-MNIST experiments were trained in the conditions described above. In each case, a random subset of $10,000$ samples is taken from the corrupted set and the TAE and MVAE models are trained with it. A random subset of $1,000$ of these is selected and ground-truth lables for these samples are made available. 

A classifier consisting in a single fully connected layer with leaky ReLu non-linearity is trained to perform classification on this subset. For each stochastic training iteration of this classifier, we generate samples associated with the corrupted observations and provide the associated labels. After the classifier is trained, we test classification performance on the remaining $9,000$ examples, by running the train classifier $400$ times per sample, each time generating clean data from a corrupted observation with the TAE and the MVAE. The histograms shown in figure 5 are built by aggregating the resulting classification. 

The above procedure is repeated $15$ times. The resulting means and standard deviations of the tested classification performance are shown in figure 6. 

\textbf{Training Conditions:} Hyper-parameters of optimisation for the models were cross validated with the MNIST data set at a proportion of missing entries of $0.9$. Hyper-parameters common to all models were determined by obtaining best performance with the MVAE model. Hyper- parameters specific to the TAE model were obtained by fixing the common parameters and cross validating these. The resulting optimal hyper parameters were then used in all other experiments of section 4.1 and 4.2, including those with different data sets. Common parameters are as follows: $500,000$ iterations with the ADAM optimiser in Tensorflow, an initial training of $2^{-4}$ and batch size of $20$. The hyper-parameters specific to the TAE are instead: $\gamma$ initially set to $0.01$ and then linearly increased to $1$ between $50,000$ and $100,000$ iterations, $\lambda = 2$ and $C = 10$. All experiments were performed using a TitanX GPU.

\textbf{NYU Rooms Experiments:} For these experiments, we take a subset of $3612$ depth maps from the NYU raw data set. We slightly crop these in one dimension to be $480 \times 608$ images. The convolutional TAE to obtain the results of figure $7$, was trained for $500,000$ iterations using the ADAM optimiser in Tensorflow, with a batch size of $2$ images and an initial training rate of $2 \times 10^{-3}$. For the warm up, we initially set $\gamma = 0.01$ and linearly increase it to $1$ between $50,000$ and $200,000$ iterations. For these experiments, $\lambda = 2$ and $C = 10$. As we are interested only in imputation, we set the data likelihood $p(y|x,\alpha, \beta) = p(y|x,\alpha)$ and this function simply masks out missing entries by multiplying them by zero according to the binary marsks $\alpha$. We then only infer the missing entries upon testing, giving the results shown in figure $7$.

\section{Additional Experiments}

\subsection{$C$ and $\lambda$ Cross-Validation}

\begin{figure}[h]
  \centering
  \includegraphics[width=10cm]{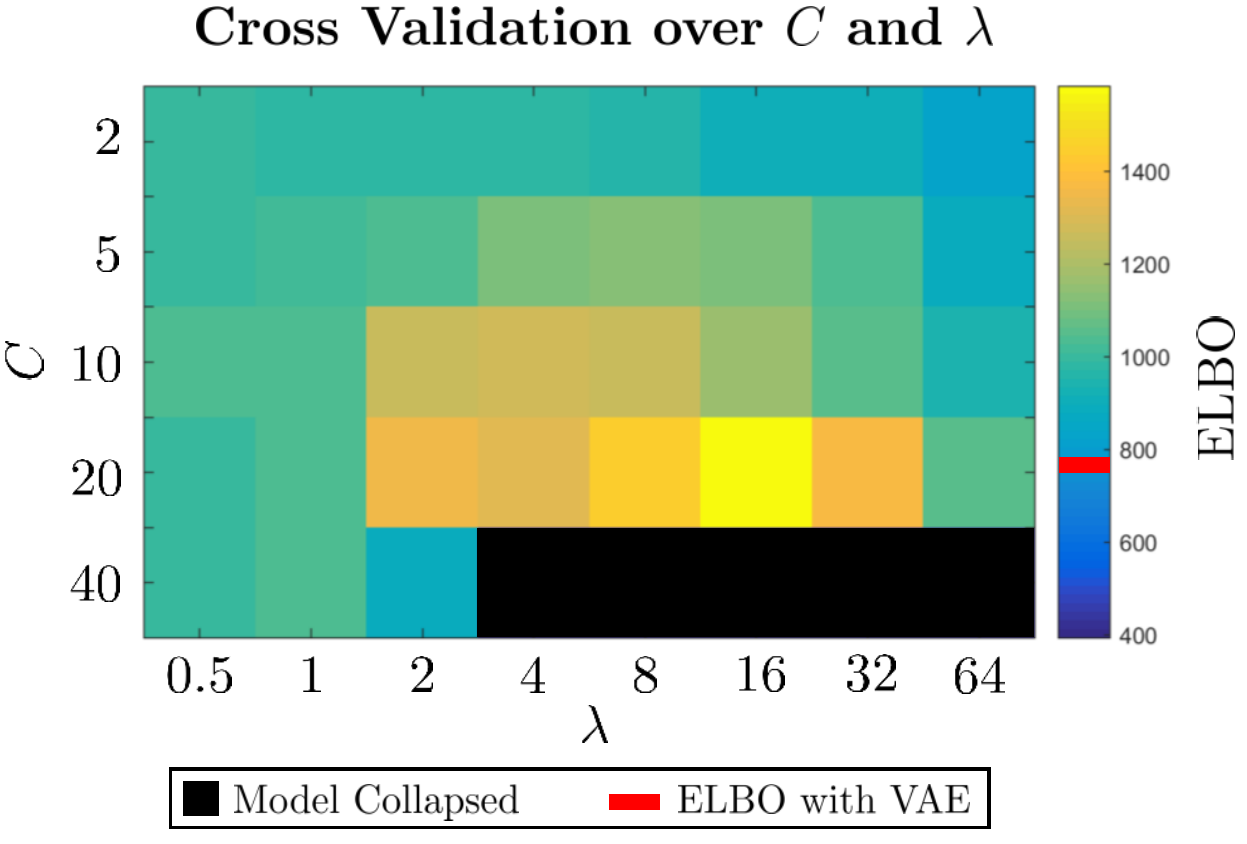}
\caption{ELBO for MNIST with $90$\% missing values and additive noise as a function of chosen hyper-parameters $C$ and $\lambda$ (in log scale). The performance of TAE exceeds that of a standard VAE approach over a broad range of values. If the values are too large, the model collapses during optimisation, making such situation easy to diagnose.}

\label{fig:cross_val}
\end{figure}

$C$ and $\lambda$ in equation \ref{TAE_OBJ} are hyper-parameters of our inference algorithm and need to be user defined. In our experiments, we determine the optimal values by cross-validation, as described in section C. We report in figure \ref{fig:cross_val} a cross validation study where we measure the TAE ELBO for MNIST with $90$\% missing values and additive noise.

As shown in figure \ref{fig:cross_val}, the performance of TAEs is robust to variations in hyper-parameters $C$ and $\lambda$ over a broad range of values. They also have an intuitive meaning that helps in their selection. In practice, $C$ controls the final value of localisation and is desirable to be as high as stability of the optimisation allows. $\lambda$ controls how fast we are imposing the model to approach $C$. 

\subsection{Missing Not-at-Random}

\begin{figure}[h]
  \centering
  \includegraphics[width=10cm]{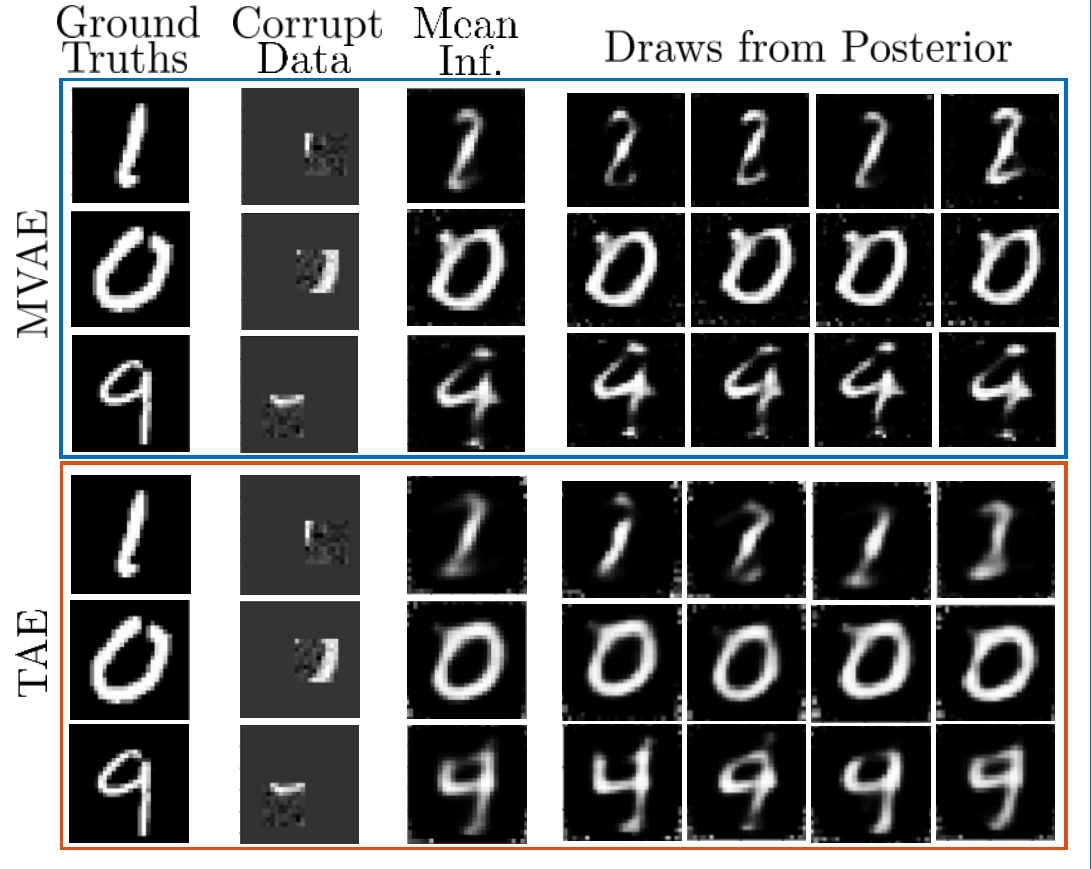}
\caption{Examples of Bayesian reconstructions with MVAE and TAE on structured missing values. the MVAE returns good mean reconstructions, but its posteriors collapse on single solutions, giving draws that are very similar to each other. The TAE returns posteriors which more broadly explore the different possible clean samples associated with the corrupted observations, giving more variation in the posterior's draws.}
\label{fig:nr}
\end{figure}

We test a TAE in a situation analogous to that shown in figure 4 of section 4, but with structured missing values instead of randomly missing ones. For each sample in MNIST, we only make visible a small $10 \times 10$ pixels window, randomly placed in each example, while the rest of the images remain hidden. In addition, the values in the observed window are subject to additive Gaussian noise, similarly to the missing-at-random case. Reconstructions with the comparative MVAE and our TAE are shown in figure \ref{fig:nr}.

Similarly to the missing-at-random case, the MVAE collapses on single solutions, giving draws from the posterior that are all very similar to each other. Contrarily, the TAE gives more variation in the possible solutions exploring more appropriately the uncertainty in the solution space. The MVAE ELBO over the clean data for this problem is $428$, while the TAE one is $638$. The performance improvement provided by the TAE is analogous to that observed with missing-at-random experiments.

\newpage

\subsection{More ELBO Evaluations}

\begin{center}
\begin{table}[h!]
\caption{ELBO assigned by the retrieved posteriors to the ground truth clean data.}
\label{tab:ELBO}
\vspace{0.3cm}
  \begin{tabular}{ l  p{3.3cm}  p{3.3cm}  p{3.3cm} }
    \toprule
     & MVAE & MIWAE & TAE \\
    \hline
    MNIST, & $883 \pm 2$ & $940 \pm 3$ & $\mathbf{1831 \pm 8}$ \\
    $20\%$ missing &   &  & \\ \hline
    MNIST, & $870 \pm 6$ & $917 \pm 4$ & $\mathbf{1719 \pm 7}$ \\
    $50\%$ missing &   &  & \\ \hline
    MNIST, & $803 \pm 15$ & $780 \pm 6$ & $\mathbf{1536 \pm 14}$ \\
    $80\%$ missing &   &  & \\ \hline
    Fashion-MNIST, & $775 \pm 4$ & $815 \pm 4$ & $\mathbf{1407 \pm 24}$ \\
    $20\%$ missing &   &  & \\ \hline
    Fashion-MNIST, & $757 \pm 1$ & $800 \pm 7$ & $\mathbf{1326 \pm 7}$ \\
    $50\%$ missing &   &  & \\ \hline
    Fashion-MNIST, & $723 \pm 7$ & $766 \pm 8$ & $\mathbf{1094 \pm 13}$ \\
    $80\%$ missing &   &  & \\ \hline
    UCI HAR, & $611 \pm 3$ & $628 \pm 10$ & $\mathbf{1039 \pm 11}$ \\
    $20\%$ missing &   &  & \\ \hline
    UCI HAR, & $585 \pm 4$ & $613 \pm 6$ & $\mathbf{1014 \pm 6}$ \\
    $50\%$ missing &   &  & \\ \hline
    UCI HAR, & $471 \pm 10$ & $584 \pm 8$ & $\mathbf{854 \pm 52}$ \\
    $80\%$ missing &   &  & \\ \bottomrule
    
  \end{tabular}
  \vspace{-0.6cm}
  \end{table}
\end{center}
\newpage
\subsection{NYU Rooms Recovery Examples}

\begin{figure}[h]
  \centering
  \includegraphics[width=10cm]{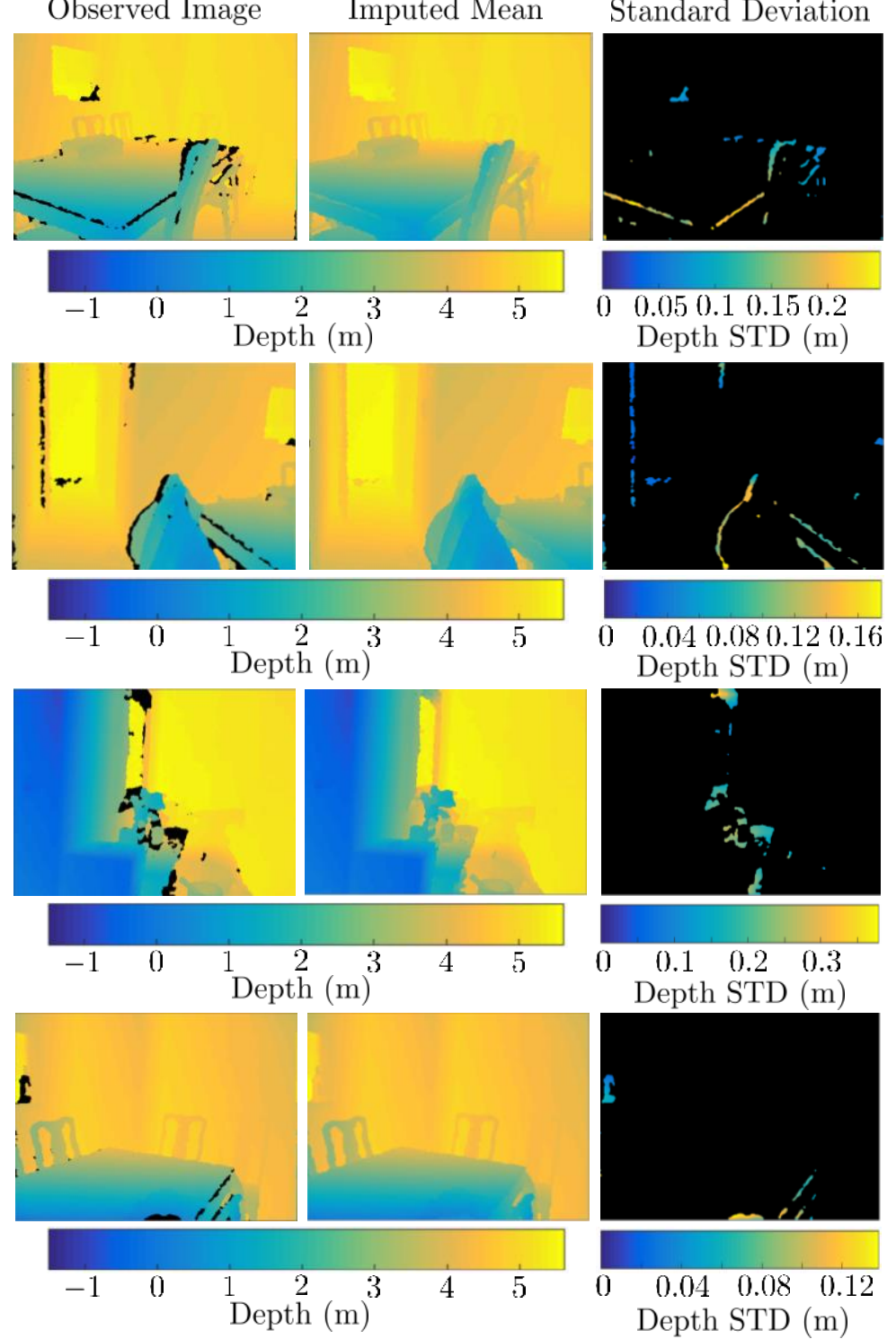}
\caption{Unsupervised missing value imputation with our TAE on raw depth maps from the NYU rooms data set. The TAE generates different possibilities for the imputed pixels, which can be aggregated to recover a mean and a standard deviation to quantify uncertainty in the retrieved imputations.}
% \vspace{-0.7cm}
\label{fig:rooms}
\end{figure}

\section*{Acknowledgements}
We  acknowledge  funding  from  Amazon  and  EPSRC  grants EP/M01326X/1, EP/T00097X/1 and EP/R018634/1.

\end{document}